# Autonomous Drone for Dynamic Smoke Plume Tracking*

Srijan Kumar Pal, Shashank Sharma, Nikil Krishnakumar, and Jiarong Hong

*Abstract*— This paper presents a novel autonomous drone-based smoke plume tracking system capable of navigating and tracking plumes in highly unsteady atmospheric conditions. The system integrates advanced hardware and software and a comprehensive simulation environment to ensure robust performance in controlled and real-world settings. The quadrotor, equipped with a high-resolution imaging system and an advanced onboard computing unit, performs precise maneuvers while accurately detecting and tracking dynamic smoke plumes under fluctuating conditions. Our software implements a two-phase flight operation: descending into the smoke plume upon detection and continuously monitoring the smoke's movement during in-plume tracking. Leveraging Proportional Integral–Derivative (PID) control and a Proximal Policy Optimization (PPO) based Deep Reinforcement Learning (DRL) controller enables adaptation to plume dynamics. Unreal Engine simulation evaluates performance under various smoke-wind scenarios, from steady flow to complex, unsteady fluctuations, showing that while the PID controller performs adequately in simpler scenarios, the DRL-based controller excels in more challenging environments. Field tests corroborate these findings. This system opens new possibilities for drone-based monitoring in areas like wildfire management and air quality assessment. The successful integration of DRL for real-time decision-making advances autonomous drone control for dynamic environments.

## I. INTRODUCTION

The atmospheric transport of Particulate Matter (PM) is an interdisciplinary field with profound implications for environmental science, climate modeling, and public health [1-3]. Examples of such transport include the dispersion of smoke plumes from forest fires, the distribution of volcanic ash during eruptions, and the movements of sand, dust, or snow migration by wind [4-7]. Understanding the dynamics of these particles is vital for predicting their environmental and health impacts, including effects on climate change, ecosystem dynamics, and respiratory health issues.

These particle transports usually span a wide range of scales, from the kilometer-scale movement of flows to the micrometer-scale size of particles [8]. Particle morphology and composition significantly influence their dispersion, yet existing field data and measurement tools are insufficient for accurately tracking these properties, which is critical for modeling [9]. Current methods like lidar and satellite imaging effectively capture large-scale particle movements but lack the resolution to provide detailed particle characteristics [8, 10], whereas in situ PM sensors, which estimate particle size distribution based on light scattering or aerodynamic properties, often rely on assumptions, leading to uncertainties, especially for irregular particles like volcanic ash [11-15].

In response to these challenges, Bristow et al. introduced an innovative autonomous drone system equipped with a Digital Inline Holography (DIH) sensor for mapping particle distribution within a smoke plume [16]. In their approach, the drone initially flies above the plume, capturing top-down images using a machine vision camera. These images are analyzed in real time using optical flow techniques to extract plume flow information, which is then used to guide the drone's navigation within the plume. Their system successfully navigated through smoke and monitored changes in particle properties during controlled experiments. However, their method's significant limitation was its reliance on top-down imaging for flow analysis, lacking intelligent navigation once the drone entered the plume. This absence of feedback control within the smoke hindered the drone's ability to adapt to shifts in wind direction, leading to inconsistent tracking in real-world scenarios characterized by rapidly changing wind patterns and turbulent environments. Therefore, we aim to overcome these limitations by developing an advanced computer vision-based control system. This system enables the drone to dynamically adjust its trajectory and respond to directional changes in the plume caused by shifting winds, thereby enhancing the performance in real scenarios.

To date, there appears to be a paucity of research specifically focusing on drones for tracking atmospheric particle transport like smoke plume dispersion. Relevant studies in this field have primarily concentrated on employing drones to track more predictable static or dynamic objects, such as vehicles, people, or other drones [17, 18]. The methods typically include tracking the motion of the target object within the camera frame [18-21] or actively following the target using the drone. Object detection is achieved in these scenarios using traditional image processing [19, 22] or deep learning approaches [23-25]. The subsequent tracking maneuvers are then executed using PID controllers [18] often coupled with Kalman filtering [20] to address uncertainties.

However, these approaches are primarily suited for well-defined and predictable objects, and they fall short when applied to unpredictable objects with complex nature of atmospheric flows like smoke. Atmospheric particle transport, such as smoke plumes or dust clouds, is fluid and dynamic,

*Research Funding Support from National Science Foundation Grant No: NSF-MRI-2018658.

Srijan Kumar Pal, Minnesota Robotics Institute, Minneapolis, MN 55455 USA and St. Anthony Falls Laboratory, Minneapolis, MN 55414 USA (phone: 763-923-4386; e-mail: pal00036@umn.edu).

Shashank Sharma, Minnesota Robotics Institute, Minneapolis, MN 55455 USA and St. Anthony Falls Laboratory, Minneapolis, MN 55414 USA (phone: 763-327-0419; e-mail: sharm964@umn.edu).

Nikil Krishnakumar, Minnesota Robotics Institute, Minneapolis, MN 55455 USA and St. Anthony Falls Laboratory, Minneapolis, MN 55414 USA (phone: 763-485-3376; e-mail: krish375@umn.edu).

Jiarong Hong, Mechanical Engineering and Minnesota Robotics Institute, Minneapolis, MN 55455, and St. Anthony Falls Laboratory, Minneapolis, MN 55414 USA (phone: 612-626-4562; e-mail: jhong@umn.edu).

Project Page: https://srijanpal07.github.io/autonomous-drone-for-dynamic-smoke-plume-tracking/

differing significantly from the more predictable objects typically tracked by drones [17, 18], [22]. This fluidity requires algorithms that adapt to continuously changing shapes, densities, and movements. Additionally, the environmental conditions where these flows occur, such as varying wind speed, direction, and turbulence, add complexity that current tracking systems, optimized for controlled settings [18, 26], struggle to handle. The current detection algorithms used on drones often produce bounding boxes that deviate from the actual centroid of the dynamic plume, leading to inaccurate tracking. Moreover, higher inference times cause delays, resulting in slow or reactive drone responses during rapid shifts in smoke flow. These challenges underscore the need for a real-time adaptive tracking solution capable of managing the unpredictability of atmospheric flows.

Recent advancements in DRL-based drone navigation have been explored to enhance adaptability and robustness in dynamic and unpredictable environments [27]. These methods include vision and depth-based localization and navigation, which are primarily applied to object avoidance, tracking, and drone racing applications [26, 28-30]. Despite the potential of DRL-based techniques, no prior research has focused specifically on using these methods to track and follow atmospheric flows, such as smoke plumes. Adapting DRL-based drone navigation to atmospheric flow tracking presents unique challenges, mainly because existing methods designed for object tracking do not adequately address the complexities of atmospheric particle transport.

To address these gaps, our study proposes a novel approach to integrating active deep learning, computer vision, and advanced control strategies when the drone gets inside the smoke plume. This enables the drone to adjust its trajectory within the plume, targeting more concentrated areas despite dynamic changes in wind conditions. This approach seeks to enhance the robustness of autonomous drone systems in tracking realistic and constantly deforming atmospheric flows in real-time based on their changing characteristics and the surrounding environmental conditions. By integrating real-time environmental data into tracking, our system seeks to achieve a level of adaptability and precision currently lacking in drone-based tracking technologies.

The structure of this paper is as follows: Section II describes the proposed drone system in detail. In Sections III and IV, we demonstrate the effectiveness of our approach through simulation and real-world field deployments. Finally, we summarize our findings and discuss their implications.

## II. METHODOLOGY

### A. Overview

Our autonomous drone-based smoke tracking system operates on a quadrotor platform equipped with a machine vision camera and an edge computing device for real-time processing, supporting tasks such as YOLO-based [31] smoke detection and control algorithms using PID/DRL controllers. As depicted in Figure 1, the system operates in two key phases: the descending phase and the in-plume tracking phase. In the descending phase, the drone begins by positioning itself above the smoke plume, using the YOLO-based smoke detection algorithm to identify the initial presence of smoke, and then initiates its descent. A PID controller governs the drone's trajectory, ensuring stable movement towards the smoke plume dispersion region, while optical flow aids in aligning the drone with the direction of the smoke flow. Upon reaching the plume, the system transitions into the in-plume tracking phase, triggered by detecting the smoke segment. During this phase, the camera reorients to maintain a continuous view of the smoke and uses YOLO-based segmentation to localize smoke in the frame. The drone's trajectory is adjusted by a combination of a PID and a PPO-based DRL controller, ensuring that the drone remains within smoke even as wind shifts and the plume changes direction, allowing the drone to continuously track the densest regions of the smoke.

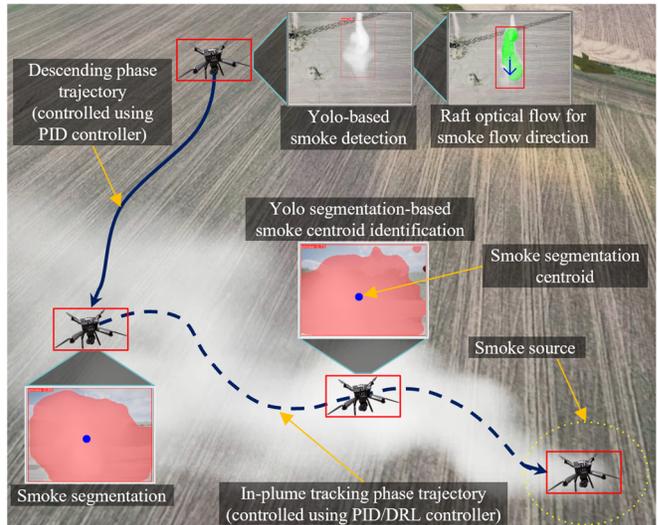

Figure 1. Autonomous drone-based smoke tracking system working principle.

### B. Hardware

The drone hardware consists of a quadrotor body, a high-resolution imaging system, and an onboard computing unit, as illustrated in Figure 2. The system builds on our previous design detailed in [16], incorporating upgrades to the imaging system and the onboard edge computing device to enhance real-time autonomy. Specifically, the core computational unit has been upgraded to the Nvidia Jetson Orin Nano, delivering up to 40 TOPS—nearly double the performance of the previous Jetson Xavier NX. It operates on Linux (Jetpack 5.1.3, Ubuntu 20.04 LTS, Linux Kernel 5.10) with ROS Noetic for efficient communication, boots from an NVMe SSD, and utilizes TensorRT to accelerate deep learning inference. To handle the high memory demands of DRL tasks, an additional 8GB of swap space has been added to the existing 8GB of RAM. The flight controller has been upgraded to the Pixhawk 6C (from Pixhawk 4), which interfaces with the Jetson through MAVROS for MAVLink communication and integrates RTK technology with GPS to achieve centimeter-level positioning accuracy. The machine vision system has been enhanced with a 12-megapixel ArduCam, replacing the previous GoPro camera to minimize latency and maintain high resolution with minimal distortion. The camera, mounted on a gimbal for both top-down and in-plume views, operates at a sensor size of 640 x 480 pixels at 30 frames per second. These upgrades significantly enhance real-time processing and autonomous control, improving the drone's multi-phase and multi-modal smoke-tracking capabilities.

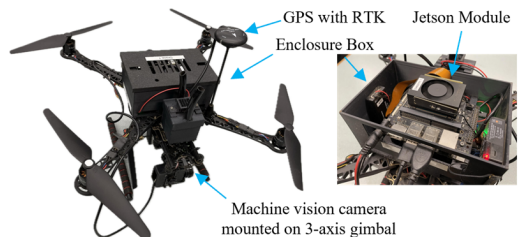

Figure 2. Autonomous drone-based smoke tracking system hardware.

*C. Algorithm and Software Architecture*

As depicted in Figure 3, the framework of the autonomous drone operation algorithm is divided into two main phases: the descending phase and the in-plume tracking phase, each comprising several steps. Precisely, the descending phase consists of the following steps:

**1) Hovering and Smoke Detection Setup:** The drone hovers above the smoke plume, with the state set to 'GUIDED' to enable autonomous controls from the Jetson module, and the gimbal pitches down the camera to capture a top-down view.

**2) Smoke Detection:** A detection node processes the top-down images using a custom-trained YOLOv8m model to detect the smoke plume with bounding boxes.

**3) Optical Flow Analysis:** Once smoke is detected, the drone halts all movement, and RAFT Optical Flow [32] analysis begins. The optical flow node processes the sequential bounding boxes to compute the smoke flow's directional vector until the mean direction gets computed.

**4) Yaw Alignment and Descent:** The drone yaws to align against the smoke flow and initiates its descent. A PID controller maintains the smoke bounding box in the upper half of the frame, ensuring the descent within the smoke dispersion area. The drone descends until it is entirely inside the smoke.

Once the drone reaches the target altitude, it switches to the in-plume tracking phase, adjusting the gimbal for a forward view from within the plume toward the smoke source. The in-plume tracking phase proceeds using the following steps:

**1) Smoke Segmentation:** At this stage, the detection node is disabled, and a smoke segmentation node takes over. The node uses a YOLOv8m-seg model trained to identify and segment denser regions within the smoke plume. It validates segments that exceed a set threshold and calculates the centroid of each valid segment by averaging the $x$ and $y$ coordinates, representing the densest smoke region. This data, including the segmentation area, centroid location, and mask, is used as input for drone trajectory control.

**2) Trajectory Control:** The drone's movement is adjusted by either a PID or DRL controller. The PID controller calculates the positional error between the smoke centroid and the center of the camera frame, generating velocity commands to adjust the drone's position along the horizontal and vertical axes. In contrast, the DRL controller processes the binary smoke segmentation mask through an actor-critic policy network, selecting actions from a discrete set of actions to maneuver the drone. The main components shown in Figure 3 operate at varying frequencies - the YOLOv8-based smoke segmentation runs at 30 Hz, performing real-time segmentation of incoming camera frames. The DRL policy inference, which determines actions based on binary smoke segmentation masks, operates at 10 Hz. Lastly, the action execution, which sends velocity commands to the flight controller via MAVROS, runs at 20 Hz to execute the actions inferred by PID or DRL controller. While the PID provides faster, more reliable responses when well-tuned, the DRL controller excels in dynamic or unpredictable environments. Currently, the system allows manual switching between the two controllers, but future versions will incorporate auto switching based on real-time smoke conditions to optimize performance. Further details on DRL customization are discussed in the following sections.

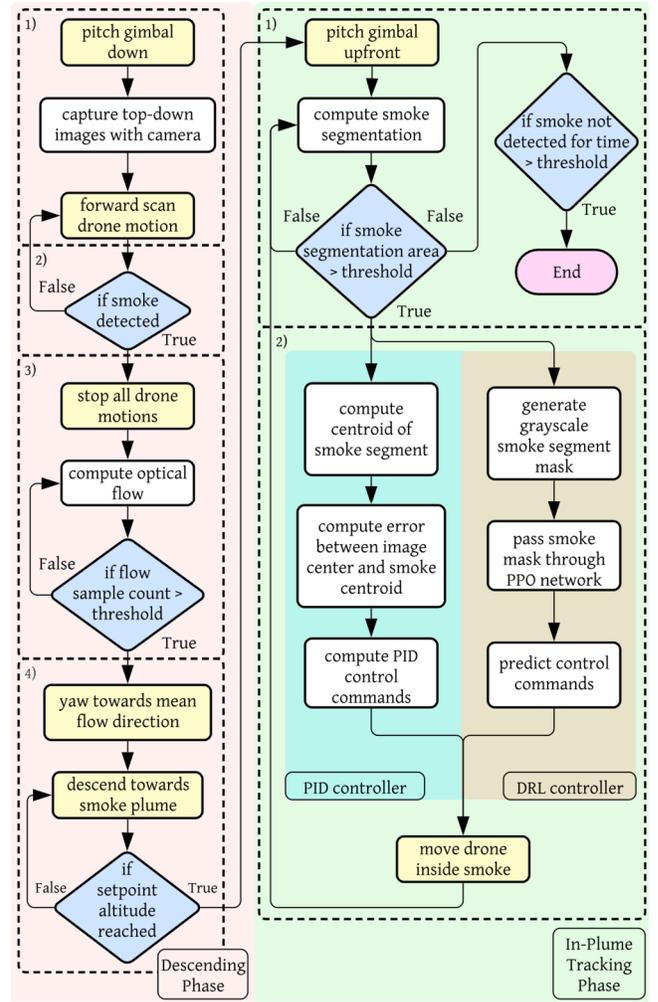

Figure 3. The framework of the autonomous drone operation algorithm.

*D. DRL Drone Control Algorithm*

Our study employs PPO, a widely used DRL algorithm implemented using the Stable Baselines3 library and OpenAI Gym environments. PPO is known for its stability, robustness in high-dimensional and stochastic environments, making it will-suited for the unpredictable nature of smoke plume dynamics. It operates using an actor-critic framework, a policy gradient method, where the actor proposes action probabilities, and the critic evaluates their expected value. PPO employs a clipped objective function to ensure stable learning, preventing extensive, destabilizing policy updates crucial for highly dynamic environments like smoke. PPO has been widely applied in various drone control tasks, such as autonomous maneuvering [33], drone tracking [34], and path planning [35]. However, adapting this approach to the unique challenges of smoke plume tracking required several adaptations.

**1) State of the Agent (Drone):** Unlike most DRL-based drone control, smoke plume tracking presents unique challenges: the agent's kinematic state (e.g., velocity, body rates, etc.) relative to the smoke flow cannot be accurately measured. Therefore, we defined the agent's state solely using single-channel binary images generated from the smoke segmentation model, as shown in the "Smoke Segmentation Mask" window in the top-right of Figure 4. These images feature white regions representing the smoke and black regions corresponding to the surrounding environment. This formulation effectively captures the location and shape of the smoke relative to the drone. Additionally, the variation in the smoke's dynamic location and shape, captured across consecutive frames, is used to train the policy. Thus, the policy learns from both the spatial and temporal information regarding the shape of smoke. This approach provides sufficient state information of the agent to effectively track the smoke plume, maximizing rewards and eliminating the need for the kinematic states.

**2) Convolutional Neural Network (CNN) Architecture for Actor-Critic Framework:** The backbone of our DRL controller is a CNN integrated into the actor-critic network of the PPO framework, designed to handle the unique visual features of smoke plumes- input layer to process the single-channel smoke segmentation masks of size 320x320 pixels, three convolutional layers with increasing filter depths (32, 64, 128) and strides to reduce spatial dimensions, followed by ReLU activation functions, a flattening layer converting the 2D feature maps into a 1D feature vector, and finally, a fully connected layer outputs a fixed-size feature vector that is fed into both the policy (actor) and value (critic) networks. This architecture is optimized for efficient feature extraction from the binary smoke masks with low latency, enabling real-time operation on the resource-constrained Jetson platform.

**3) Discrete Action Space:** The DRL predicts within a discrete action space with seven movement options [0-6], each corresponding to predefined velocities $V_x$, $V_y$, and $V_z$ (in m/s) along the drone's horizontal (y) and vertical (z) axes with constant velocity along the forward direction (x): [0] Up ($V_y$=0, $V_z$ >0); [1] Hard left ($V_y$=−$mV_y$, $V_z$=0); [2] Left ($V_y$=−$V_y$, $V_z$=0); [3] No movement ($V_y$, $V_z$=0); [4] Right ($V_y$=$V_y$, $V_z$=0); [5] Hard right ($V_y$=$mV_y$, $V_z$=0); [6] Down ($V_y$=0, $V_z$<0), where $m$ is the multiplying factor for faster velocities and magnitudes of $V_x$, $V_y$, $V_z$ and $m$ has been empirically assigned. The range of actions, from subtle actions (left, right) to more aggressive maneuvers (hard left, hard right) offer the flexibility needed to navigate effectively inside smoke. These high-level velocity commands are published to the MAVROS topic "/mavros/setpoint_velocity/cmd_vel" to communicate with the flight controllers that compute individual motor signals needed to achieve the targeted velocities.

**4) Reward Function:** The reward function is formulated to incentivize the drone to track denser smoke by positively rewarding actions that make the drone move towards the smoke. The reward is based on the smoke's location in the image, which is divided into seven regions, and the DRL controller's prediction [0-6], which corresponds to the seven image regions and is described in detail in Figure 5.

**5) Data Collection for Smoke Segmentation:** The smoke segmentation model is trained on simulated and real-world smoke data for generalization. To account for real-world ambient lighting variations, we collected smoke images at different times and under various weather conditions, including bright, sunny, and cloudy days. The dataset includes over 2,000 manually annotated images. Despite this, we found that fine-tuning the camera settings (e.g., exposure time, saturation, etc.) was necessary before each deployment to ensure accurate smoke segmentation. Currently, the model is trained on white smoke images; future work will include smoke of different colors to improve generalization.

**6) DRL-Controller Training Configuration:** To ensure robustness, we simulated smoke with realistic disturbances in Unreal Engine by varying wind speed and direction. During training, the wind conditions randomly alternated between steady linear and highly fluctuating flows with different oscillation frequencies, allowing the agent to adapt to various real-world conditions and improve performance.

The DRL controller was trained with the following hyperparameters to promote efficient learning and stability: a learning rate of $3×10^{−4}$, a discount factor $\gamma = 0.99$, a batch size of 256, 2048 steps per update, and 10 epochs per update, with a total training duration of 1 million time steps. This configuration ensures adaptability and stable learning.

**7) Inference and Simulation-to-Real (Sim2Real) Policy Transfer:** During inference, the trained model applies the learned policy to control the drone's movements in simulated and real-world smoke environments. The segmentation model, trained on actual and simulated smoke, generates binary smoke masks. The DRL controller then uses these binary smoke masks to predict drone actions mapped to drone control commands, as discussed in section II.D.3. Since the segmentation model handles both the simulated and actual smoke images and the DRL controller functions solely on the binary smoke segmentation masks, this approach helps in seamless policy transfer from simulation to real-world and results in effective drone navigation across both environments.

## III. SIMULATION ASSESSMENT

### A. Simulation Environment

Developing an autonomous drone system for smoke plume tracking presents several challenges, including the need for large, open testing areas, unpredictable drone behavior that can lead to costly crashes, and weather conditions like wind and rain. We developed a simulation environment using Unreal

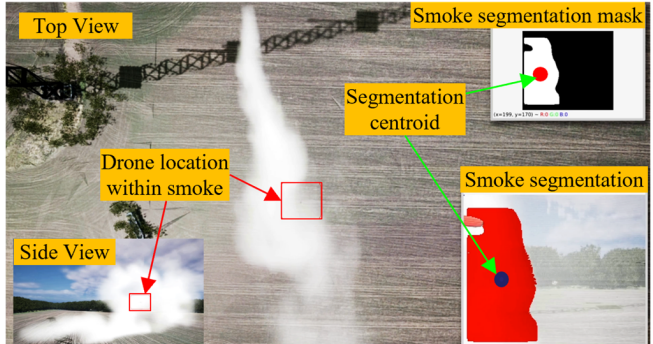

Figure 4. Simulation of the deployment of autonomous drone-based smoke tracking system at Eolos Field Station, Rosemount, Minnesota, USA.

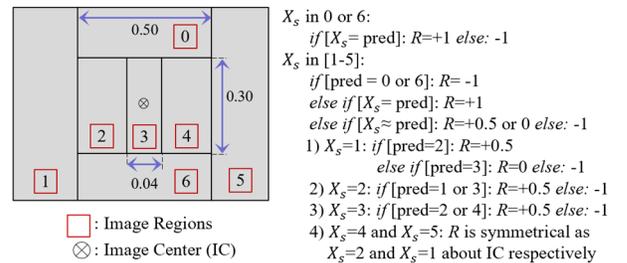

Figure 5. Illustration of the reward function, where $X_s$ is the smoke centroid location in image, *pred* is the DRL controller's prediction, and *R* is the reward. The camera image is partitioned into 7 regions symmetrical about the image center (IC), with the dimensions indicating fractions of the total image length.

Engine 5.1.1 to overcome these challenges and enable rapid algorithm testing and refinement under controlled, realistic smoke-wind scenarios. The simulation replicates real-world conditions at our primary field-testing site, the Eolos Field Station in Rosemount, Minnesota, USA. The simulation includes terrain, vegetation, and infrastructure to create an environment that mirrors the testing site closely. For smoke generation, we utilized Unreal Engine's Niagara Plugin, which includes Niagara Fluids, Chaos Niagara, and the Niagara Custom Data Interface, to produce realistic smoke dynamics. The setup allows complete control over wind speed and direction through a custom blueprint in the event graph, which generates time-dependent wind vectors supporting variable and constant direction wind flows, allowing flexibility in test conditions. The drone's flight control is simulated using the PX4 Software-in-the-Loop (SITL) integrated with AirSim, which simulates drone sensors. These data are published on ROS topics in Windows Subsystem for Linux (WSL2) emulating a virtual Linux environment similar to a drone with Jetson. The MAVROS package ensures seamless MAVLink-based communication with PX4 SITL for autonomous drone control. Overall, this simulation provides a robust platform for training, algorithm feasibility testing, and smoke detection under a wide range of realistic smoke-wind scenarios. The flexibility and reproducibility of the Unreal Engine smoke simulation allow us to evaluate the drone's performance in diverse test conditions before field deployment.

## B. Performance Assessment Using Simulation

The smoke tracking performance of the drone, using both PID and DRL-based control algorithms, was evaluated under different smoke generation and wind conditions in the simulation. Four test scenarios, with increasing levels of tracking difficulty, were designed to assess the algorithms:

**1) Steady Smoke Flow (S)**: The smoke plume is steady under constant streamwise wind, $V_{w,y}$ of 4.5 m/s with no fluctuations in direction or speed. This represents a stable and predictable condition, ideal for baseline performance evaluation.

**2) Unsteady Smoke Flow with Low-Frequency Horizontal Fluctuation (UL):** The smoke experiences a mild low-frequency fluctuating crosswind, which is superimposed on top of the primary wind of $V_{w,y} = 4.5$ m/s. The crosswind is specified as $V_{w,x} = 1.35 \sin(0.02\pi t)$ m/s with an amplitude of 1.35 m/s and a frequency of 0.01 Hz.

**3) Unsteady Smoke Flow with High-Frequency Horizontal Fluctuation (UH):** The smoke experiences a stronger high-frequency crosswind superimposed on top of the primary wind $V_{w,y} = 4.5$ m/s. The crosswind is specified with $V_{w,x} = 1.95 \sin(0.04\pi t)$ having an amplitude of 1.95 m/s and 0.02 Hz frequency. Under such conditions, the smoke changes direction more rapidly, increasing tracking challenge.

**4) Unsteady Smoke Flow with 3D Fluctuation (U3D):** Both horizontal and vertical wind fluctuations are introduced on top of the primary wind $V_{w,y} = 4.5$ m/s to increase the tracking challenges further. The horizontal crosswind is specified as $V_{w,x} = 1.95 \sin(0.04\pi t)$ (amplitude 1.95 m/s, frequency 0.02 Hz), and vertical wind $V_{w,z} = 0.3 \sin(0.02\pi t)$ (amplitude 0.3 m/s, frequency 0.01 Hz). Under these, the plume fluctuates in a complex 3D pattern.

To accurately assess the drone's smoke tracking performance, we positioned an observer drone equipped with a camera at a fixed altitude above the smoke plume, providing a top-down view of the tracking drone, as shown in Figure 6a. As the tracking drone enters the smoke plume, visual methods like bounding boxes become ineffective. Instead, we mapped the tracker drone's 3D GPS coordinates to 2D-pixel coordinates within the observer drone's camera image. Using the observer drone's GPS location and camera focal length and applying the haversine formula, we computed the image's dimensions in the world frame. We used an affine transformation matrix to map each pixel to GPS coordinates. The inverse transformation allowed us to accurately track the drone's location in the camera frame without relying on image-based detection techniques.

As shown in Figure 6b, custom metrics are introduced to compare PID and DRL controllers based on the tracking drone's position relative to the smoke plume contour to evaluate how close the drone stays within the plume center during the entire tracking duration. The plume contour is generated through thresholding and extracting the largest contour, with the red circle representing the drone's location and the green spline depicting the smoke plume's mean line (skeleton). In total the following five metrics are used for performance evaluation including normalized average distance of the drone from the mean line $\tilde{\mu}_m = mean(d_m) / L_{ref}$, normalized maximum distance of the drone from the mean line $\tilde{d}_{m,max} = (d_m)/L_{ref}$, normalized average distance when outside the smoke plume $\tilde{\mu}_c = mean(d_c) / L_{ref}$, normalized maximum distance when outside the smoke plume $\tilde{d}_{c,max} = (d_c)/L_{ref}$, percentage of time inside the smoke plume $\tilde{t}_R$, where $L_{ref}$ is the total smoke tracking length, $d_m$ is the drone's distance from the smoke mean line, and $d_c$ is the distance from the smoke contour.

Table 1 summarizes the performance metrics for both PID and DRL controllers under different test conditions. In steady smoke flow (S), both PID and DRL controllers perform similarly, effectively tracking the plume with comparable $\tilde{\mu}_m$ and $\tilde{t}_R$, indicating that both controllers can track the steady smoke plume effectively. Under low-frequency, low-amplitude unsteadiness (**UL**), the DRL controller shows a marked improvement in tracking the mean line of the plume. The $\tilde{\mu}_m$ drops significantly from 4.0% (PID) to 1.8% (DRL), highlighting its ability to maintain closer alignment with the plume. The variability remains comparable, indicating that DRL's performance gain is not due to increased instability but improved trajectory correction. In higher-frequency unsteady flow (**UH**) with stronger crosswinds, the DRL outperforms PID across all metrics, notably improving in $\tilde{d}_{c,max}$, where the DRL controller reduces the deviation to 12.3%, compared to 18.6% for PID, and significantly extending the $\tilde{t}_R$ achieving 85% compared to 69% for PID. Finally, under 3D unsteadiness (**U3D**), the DRL controller exhibits significant improvements across all metrics, showing a much closer alignment with the smoke, $\tilde{\mu}_m$ significantly reduced compared to PID, and achieves a greater $\tilde{t}_R$. Overall, the DRL

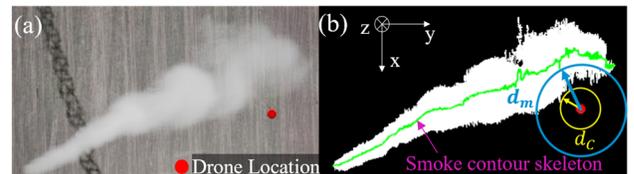

Figure 6. Illustration of (a) top-view images and (b) the metrics used to evaluate the performance of our drone-based smoke tracking system.

controller demonstrates superior adaptability in this highly dynamic and complex smoke environment.

|  |  | $\tilde{\mu}_m$(%) | $\tilde{d}_{m,max}$(%) | $\tilde{\mu}_c$(%) | $\tilde{d}_{c,max}$(%) | $\tilde{t}_R$ (%) |
|---|---|---|---|---|---|---|
| S | PID | 1.6±0.2 | 7.5±2.1 | 1.4±0.6 | 2.9±0.7 | 95.1±2.5 |
| S | DRL | 1.4±0.2 | 7.0±2.3 | 1.8±1.2 | 4.0±3.3 | 94.1±1.4 |
| UL | PID | 4.0±0.4 | 11.9±2.2 | 2.5±1.3 | 6.6±3.1 | 87.1±2.2 |
| UL | DRL | 1.8±0.5 | 10.1±2.3 | 2.5±1.1 | 8.4±3.3 | 86.9±1.8 |
| UH | PID | 7.2±1.5 | 28.1±10.9 | 5.3±3.8 | 18.6±7.4 | 69.4±4.7 |
| UH | DRL | 5.4±1.1 | 26.0±8.4 | 4.6±3.1 | 12.3±6.5 | 85.0±6.4 |
| U3D | PID | 4.1±0.6 | 15.5±6.5 | 4.5±2.6 | 12.8±5.9 | 79.5±2.3 |
| U3D | DRL | 1.9±0.5 | 10.6±4.4 | 1.5±2.1 | 4.3±4.3 | 95.0±4.1 |

Table 1. Performance metrics were evaluated using the simulation, and mean and standard deviation were calculated based on five tests under each condition for both PID and DRL controller separately.

## IV. FIELD DEMONSTRATION

Field testing was conducted at the Eolos Field Station at Rosemount, Minnesota, USA, an open agricultural field ideal for controlled experiments. Non-harmful smoke plumes were generated using a high-density smoke generator that combined food-grade chemicals (glycerin, propylene glycol, and artificial smoke fluids) in proportions, producing the plume shown in Figure 7. Deployments were performed under S-SE winds, with speeds of 4.9-6.7 m/s (11–15 mph), gusts up to 8.8m/s, and temperatures around 25°C, ensuring optimal visibility and safe drone operations. Before each experiment, all scripts for smoke tracking were initialized (image capture, smoke segmentation, and drone control with either PID or DRL), and the drone is set to 'GUIDED' mode to respond to autonomous control commands only when the smoke plume was established. The test began with the drone hovering near the plume's dispersing region, after which it autonomously tracked the smoke by adjusting its trajectory in response to wind direction changes. Experiments were conducted using both PID and DRL control algorithms in separate tests. In both cases, the drone consistently maintained its track as the plume shifted, dynamically adjusting its trajectory to follow it and eventually reach the smoke source. As shown in Table 2, the field performance metrics indicate that the DRL controller performs better than the PID controller in staying inside the smoke plume for a more extended period, achieving a $\tilde{t}_R$ of 72.7% compared to 71.2% for PID. However, the PID controller maintained closer proximity to the smoke plume core, with $\tilde{\mu}_m$=6.4 compared to $\tilde{\mu}_m$=8.1 for the DRL controller. The metrics $\tilde{d}_{c,max}$ and $\tilde{\mu}_c$ also suggest that PID remained slightly closer to the smoke plume than DRL. However, unlike in the simulation, it was not possible to maintain identical smoke conditions during the PID and DRL experiments in the field, meaning the results may not quantitatively mirror the trends observed in the simulation.

|  | $\tilde{\mu}_m$(%) | $\tilde{d}_{m,max}$(%) | $\tilde{\mu}_c$(%) | $\tilde{d}_{c,max}$(%) | $\tilde{t}_R$(%) |
|---|---|---|---|---|---|
| PID | 6.4 | 21.0 | 3.6 | 10.0 | 71.2 |
| DRL | 8.1 | 27.8 | 4.1 | 12.2 | 72.7 |

Table 2. Performance Metrics evaluated in field deployment

## V. CONCLUSION AND DISCUSSION

This paper presents an advanced autonomous drone-based smoke plume tracking system capable of navigating and tracking plumes in highly unsteady atmospheric conditions. The system integrates sophisticated hardware and software in a quadrotor platform with imaging systems and an onboard computing unit. The drone tracks dynamic smoke plumes using a combination of PID and DRL controllers. Our method implements a two-phase flight operation: the descending phase, during which the drone descends into the smoke plume, and the in-plume tracking phase, during which it continuously tracks the smoke's movement. Simulations and field tests demonstrate that while the PID performs adequately in simpler scenarios, the DRL-based controller excels in challenging environments with high fluctuations. Field tests corroborated the effectiveness of our approach, yielding results consistent with those observed in the simulation. This system significantly improves the ability to autonomously track smoke plumes in realistic atmospheric environments, providing a substantial advancement over previous methods focused on static or predictable objects.

This work builds on our previous study by enabling autonomous tracking of dynamic smoke plumes and enhancing real-time control through deep learning and DRL. The system has broad potential applications beyond smoke tracking, including wildfire monitoring, air quality assessment, environmental hazard tracking, and particle transport studies. Its ability to track dynamic plumes could benefit for emergency response operations and study atmospheric phenomena like fog, pollution clouds, and volcanic ash plumes. The successful integration of DRL for real-time decision-making in complex environments represents a major step forward in autonomous drone control.

Despite promising results, several limitations persist. The smoke detection and segmentation model, trained on specific datasets, may not generalize effectively to all smokes or atmospheric conditions. Furthermore, more rigorous testing in harsher and more unpredictable conditions is required to validate the system's robustness. Future work will focus on improving the generalization of the smoke detection and segmentation models, enhancing the DRL controller to handle more extreme environmental changes, and expanding the system's capabilities for fully autonomous operation with minimal manual intervention. We will also explore extending this approach to other atmospheric and environmental applications, such as tracking fog, pollution, or ash plumes.

## ACKNOWLEDGMENT

This work was supported by the NSF Major Research Instrumentation program (NSF-MRI-2018658), which funded the development of the drone system. We would like to sincerely thank Subhas C. Pal, Mayan Iyer, Sujeendra Ramesh, S.G.L.A. Divakarla, and Rammesh A. Saravanan for their invaluable assistance during deployments. We also acknowledge the use of ChatGPT to refine this manuscript.

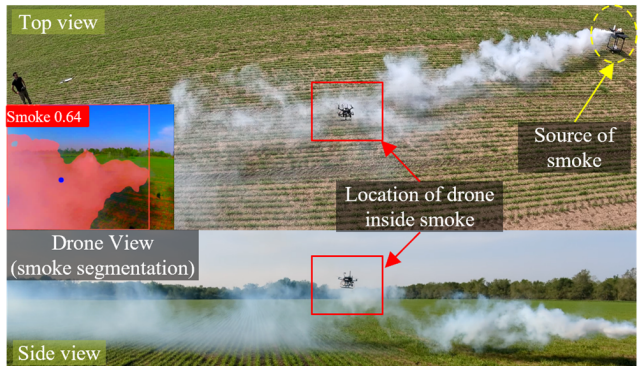

Figure 7. Field demonstration of autonomous drone-based smoke tracking algorithm at Eolos Field Station, Rosemount, Minnesota, USA.